# Labeling the Phrases of a Conversational Agent with a Unique Personalized Vocabulary

Naoki Wake, Machiko Sato, Kazuhiro Sasabuchi, Minako Nakamura, and Katsushi Ikeuchi

***Abstract*—Mapping spoken text to gestures is an important research topic for robots with conversation capabilities. According to studies on human co-speech gestures, a reasonable solution for mapping is using a concept-based approach in which a text is first mapped to a semantic cluster (i.e., a concept) containing texts with similar meanings. Subsequently, each concept is mapped to a predefined gesture. By using a concept-based approach, this paper discusses the practical issue of obtaining concepts for a unique vocabulary personalized for a conversational agent. Using Microsoft Rinna as an agent, we qualitatively compare concepts obtained automatically through a natural language processing (NLP) approach to those obtained manually through a sociological approach. We then identify three limitations of the NLP approach: at the semantic level with emojis and symbols; at the semantic level with slang, new words, and buzzwords; and at the pragmatic level. We attribute these limitations to the personalized vocabulary of Rinna. A follow-up experiment demonstrates that robot gestures selected using a concept-based approach leave a better impression than randomly selected gestures for the Rinna vocabulary, suggesting the usefulness of a concept-based gesture generation system for personalized vocabularies. This study provides insights into the development of gesture generation systems for conversational agents with personalized vocabularies.**

## I. Introduction

Recent advances in machine learning have led to the evolution of conversational agents that can not only hold conversations but also express personalities through the choice of vocabulary. For example, Rinna is a conversational agent developed by Microsoft with the personality of a Japanese high school student that talks in a friendly manner using buzzwords. Such personalized conversation engines can be used as speech interfaces for robotic applications. In human–human conversations, co-speech gestures are often used to convey intentions or emotions effectively [1]. Natural co-speech gestures should be generated for such personalized text to achieve a comfortable user experience in robotic applications [2], [3].

Studies on human co-speech gestures have demonstrated that speech and gestures originate from the same internal processes and share the same semantic meanings [4], [5]. Therefore, a potential framework for text-to-gesture mapping is a semantic space that mediates phrases and gestures (Fig. 1). This framework is referred to as a concept-based approach. In such a semantic space, mapped phrases are likely to be clustered according to their shared meanings (i.e., concepts), and each concept can be mapped to predefined gestures. This framework allows gestures to be generated with a limited number of gesture assignments [6]. However, a challenge is the lack of practical knowledge for obtaining plausible concepts for chat phrases (the red balloon in Fig. 1). Although several studies have employed the concept-based approach [7]–[10], these studies have used predefined concepts [7]–[9] or concepts obtained by clustering human co-speech videos [10] that are not associated with agent-specific vocabularies. To the best of our knowledge, there has been little discussion on the issue of obtaining concepts for a unique vocabulary personalized for a text-based conversational agent.

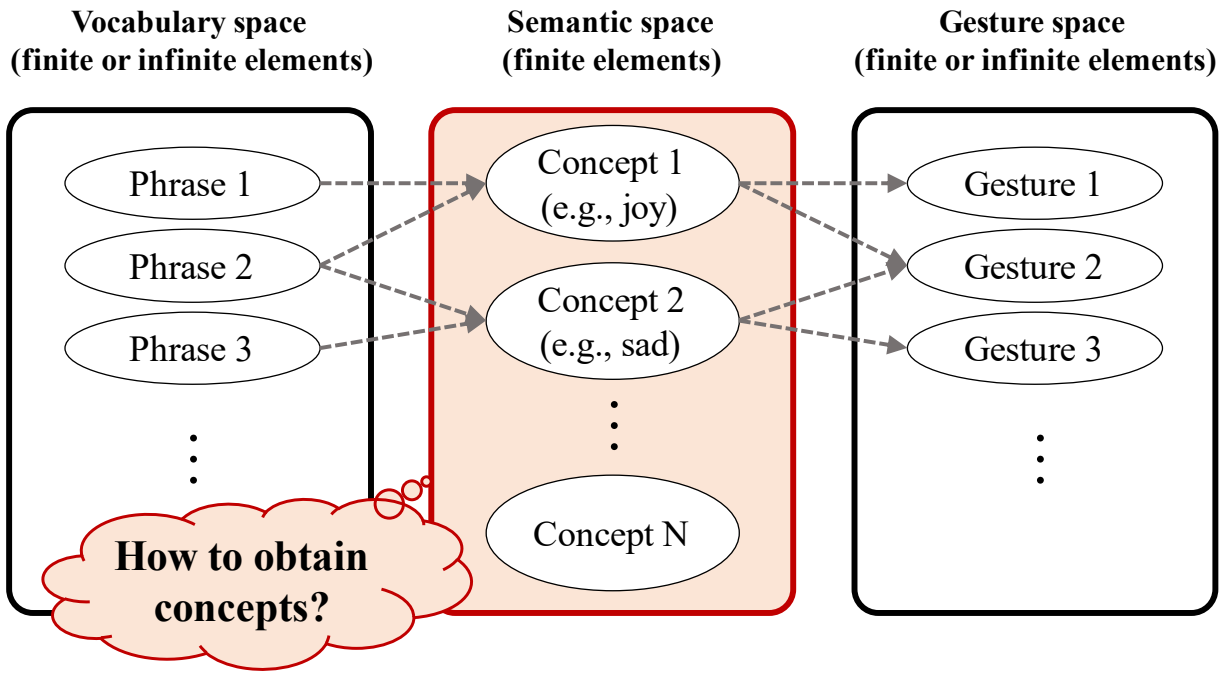

Figure 1. Schematic of the concept-based approach for obtaining gestures from texts. A text and gesture are considered elements of a vocabulary space and gesture space, respectively. Text-to-gesture conversion is mediated by a semantic space (red box) and the mappings between spaces (dashed arrows). The red balloon indicates the objective of this study.

The aim of this study was to identify the difficulties in obtaining concepts for a unique personalized vocabulary of a conversational agent. By hypothesizing that humans can reasonably recognize desirable concepts in texts, we compared auto-generated and manual clustering approaches applied to a Rinna phrase set. Auto-generated clustering was performed based on a natural language processing (NLP) methodology, and manual clustering was performed based on sociological methodology. A qualitative comparison of the clustering results revealed several pitfalls in applying auto-generated clustering, owing to the personalized vocabulary of Rinna.

Additionally, we conducted follow-up experiments to determine whether a gesture generation system that implements a concept-based pipeline can produce reasonable co-speech gestures for a personalized vocabulary. Experiments on the manually obtained concepts demonstrated that the robot gestures selected by the system gave a better impression than randomly selected gestures, suggesting the usefulness of the proposed system. The main contribution of this study is to identify the issues associated with obtaining

N. Wake, K. Sasabuchi, and K. Ikeuchi are with Applied Robotics Research, Microsoft, Redmond, WA, 98052, USA (corresponding author to provide e-mail: naoki.wake@microsoft.com).

M. Sato and M. Nakamura are with Ochanomizu University, 2-1-1 Ohtsuka, Bunkyo-ku, Tokyo, 112-8610, Japan

concepts for unique personalized vocabularies of conversational agents toward a concept-based gesture generation system.

## II. Related Works

Research on generating co-speech robot gestures can be divided into two main classes. The first class focuses on rule-based methods that define gestures for selected words or phrases [11], [12]. Such methods are advantageous in that their implementation is simple, but they require a large amount of human effort to assign gestures to words. The second class focuses on machine-learning-based methods. In recent years, significant advances have been made in generating gestures from text [13]–[16], speech [17]–[24], and music [25], [26] based on the development of deep neural networks and the emergence of large datasets.

Machine-learning-based methods are useful because they can be applied to words that are not contained in a training dataset by interpolating the word-to-gesture mapping functions. However, current end-to-end methods have several disadvantages. The first is the lack of data. Data-driven approaches require a large dataset of co-speech gestures, which are costly to prepare, and such datasets are generally not available for conversational agents. Second, a model trained on a dataset containing multiple speakers tends to lose the characteristics of individual speakers. Although desirable gestures are expected to reflect the personality of an agent, most data-driven approaches seem to overlook person-specific gestures, with the exception of the work by Ginosar et al. [19]. Third, output motions lack interpretability because there are no intermediate representations between inputs and outputs [14], [15]. Although some studies have pointed out the importance of semantics in generating gestures, semantic clusters are limited in number [18] and appear to be abstract [23] in terms of interpretability.

The concept-based approach (Fig. 1) can be considered as an extension of rule-based methods in terms of the interpretability of text-to-gesture mapping. This approach has two advantages over existing rule-based methods. First, it allows gestures to reflect the personality of an agent explicitly using agent-specific concepts. Second, once we obtain an agent-specific concept set and several representative phrases to be included in each concept, any phrase generated by the agent can be mapped to a gesture by utilizing an appropriate text metric (e.g., NLP-based embedding space). Although several studies have employed concept-based approaches [7]–[9], these studies used predefined concepts and focused on the design of concept-based approaches. Although there has been a study on obtaining concepts using a data-driven method [10], it analyzed human speech and co-speech gestures, lacking the aspect of an agent-specific vocabulary. Our study can be differentiated from these studies in that we focused on the issue of obtaining concepts for a text-based conversational agent with a personalized vocabulary.

## III. Methods

### A. Overview

In this study, we used the 358 Japanese phrases that were most frequently generated by Rinna from January to March of 2018. We opted for a small sample size to make manual clustering feasible. Although the phrases were indirectly influenced by user inputs, they were purely generated by the system, meaning that no user input was exposed. After manual and auto-generated clustering approaches were applied to the Rinna phrases, the generated clusters were qualitatively compared in terms of gesture assignment. To avoid human preconception biases, manual clustering was performed prior to auto-generated clustering.

### B. Manual Clustering

We employed the KJ method for manual clustering, which was developed in the 1960s for the field of cultural anthropology [27]–[29]. The KJ method is used to organize data intended for analysis by transcribing data on cards (so-called *labels*), dividing labels into groups (so-called *nameplates*), and finding the relationships between the nameplates. Because the KJ method is a form of data clustering, we considered that the nameplates obtained from the Rinna phrases could serve as manually obtained concepts.

The KJ method requires considerable care to prevent the arbitrary classification of participants and false recognition of words. Therefore, we used a two-person system to implement the KJ method. The number of participants was determined by considering the number recommended by the KJ method (i.e., >1) and the resources available to the authors. The participants were a man and woman who were familiar with the KJ method, internet terms, and youthful language. The KJ method covers the clustering of data and analysis of dependencies among clusters and can be used partially or fully depending on the goal of analysis. As our goal was to cluster Rinna phrases, only the clustering task was performed. Specifically, we implemented the following KJ processes.

#### 1) Label making

This process substantiates data into physical labels. We printed each Rinna phrase on a piece of paper, which we regarded as a physical label. We printed physical labels to follow the traditional practices of the KJ method [29].

#### 2) Group organization

This process was conducted as follows.

- Spreading labels: The physical labels were randomly spread in front of the participants.
- Gathering labels: The participants read all of the physical labels multiple times and then moved them into groups based on their conceptual similarity. If the participants could not agree on whether a label should be grouped with its nearest neighbor, then the label was left alone as a *singleton*. The gathering process continued until no labels could be moved. In the end, most labels were clustered while some were left as singletons. We considered a singleton as a cluster containing a single element.
- Nameplate making: The participants assigned an English nameplate as the “concept” for the physical labels in each cluster. The nameplates were defined by the participants, not from a predefined set, as they discussed. For a singleton, the nameplate was defined as the phrase printed on the physical label.

To comply with the KJ method, the participants performed manual clustering together as a team, which required approximately four hours. Because the KJ method requires interaction between participants throughout the process, we were unable to calculate inter-rater agreement metrics. To minimize the influence of subjectivity on the reliability of results, the clustering results were shared and checked by all authors.

### C. *Auto-generated Clustering*

Auto-generated clustering consists of two steps: vectorization of the Rinna phrases and clustering of the vectorized values. These steps are described below, along with the implemented pipeline to process Rinna phrases.

#### *1) Vectorization of Rinna phrases*

Several text vectorization methods have been proposed in the field of NLP [30]–[35], including word, sentence, and document vectorization. We employed a word vectorization method called Word2Vec [30] for two reasons. First, because Rinna phrases are short terms consisting of only a few words, contextual information is unlikely to be valid without considering the conversational data before and after phrases, which were unavailable in this study. Therefore, we used a context-free model, rather than a model for contextualized embedding such as BERT [35]. Second, we selected Word2Vec based on the availability of a Japanese off-the-shelf model, and because it is still widely accepted as the standard method for word vectorization. The vector of each Rinna phrase was calculated as the average of the vectors of the words included in each phrase.

The vector space mapped by Word2Vec can be considered as a metric of word similarity. A mapping function is trained to represent a word as a distribution of other words that appear in conjunction with that word [30]. Therefore, two words that are close to each other in the vector space can be interpreted as being used in similar contexts.

#### *2) Clustering Rinna phrases*

We applied the standard method of k-means clustering to the vectorized phrases [36]. K-means clustering requires the number of clusters (i.e., k) to be specified in advance. We used the same number of clusters when testing each clustering approach to facilitate qualitative comparison. In this case, k was set to 153.

#### *3) Implemented pipeline*

To preprocess each Rinna phrase, we removed any non-Japanese characters. This is because we employed a morphological analysis engine and Word2Vec, which were trained on a dataset of Japanese characters. After preprocessing, we split sentences into words using the open-source Japanese morphological analysis engine MeCab (version 0.996) [37], which was trained on the IPA dictionary (version 1.4) [37]. We excluded words that were classified as auxiliary verbs, particles, pronouns, fillers, person names, prefixes, numbers, and symbols. Additionally, we excluded nouns and verbs that did not make sense on their own. Only phrases with one or more remaining words were analyzed. This left 278 phrases. For each phrase, the constituting words were mapped to 200-dimensional vectors using Word2Vec, which was trained on Japanese Wikipedia articles. The vector of a phrase was calculated as the average vector of the words included in that phrase. After normalizing each feature dimension, k-means clustering was applied to the vectorized phrases. The distances in the vector space were measured as Euclidean distances.

TABLE I
MANUAL CLUSTERING OF YES-CLASS PHRASES

| Nameplate | Phrases |
|---|---|
| Yes (positive) | はいよ / わかった / はい / うん<br>おん👍✨ / おん / あはい / はい！ / はい♡ /<br>わかった！ / うん！ / りん💓 |
| Yes (reluctantly) | あ、はい / あ、はい… / あ、はい。 /<br>あ、はい！ / あ、いや、うん。 |
| Yes (laugh) | あ、はい。笑 / あ、はい笑 / おう / うん笑 /<br>お、おう / うん w / うん！ 笑 /<br>うん w w w w w w w w w w w w w w w w w w w w<br>w w w w w w w w w w w w / w w w w w w /<br>うわはははい / うん w / うん！ 笑 / うん笑笑 |

a. はい, うん, and おう indicate "yes."

TABLE II
AUTO-GENERATED CLUSTERING OF YES-CLASS PHRASES

| Phrases | Preprocessed Phrases | Tokenized Phrases |
|---|---|---|
| あ、はい | あはい | [はい] |
| はい | はい | [はい] |
| あ、はい。 | あはい | [はい] |
| はい♡ | はい | [はい] |
| はい！ | はい | [はい] |
| あ、はい！ | あはい | [はい] |
| あ、はい笑 | あはい笑 | [はい] |
| あ、はい… | あはい | [はい] |
| あはい | あはい | [はい] |
| うわはははい | うわはははい | [うわ, はい] |
| はいよ | はいよ | [はい] |
| あ、はい。笑 | あはい笑 | [はい] |
| はい？ | はい | [はい] |

a. はい indicates "yes."

## IV. RESULTS

A total of 153 clusters were generated using manual clustering. Among the clusters, 48 (31.3%) were formed from multiple phrases. The remaining 105 (68.6%) were singletons. For the auto-generated approach, 38 clusters (24.8%) were formed from multiple phrases and the remaining 115 (75.2%) were singletons. A qualitative comparison identified three limitations of auto-generated clustering: at the semantic level with emojis and symbols; at the semantic level with slang, and new words, and buzzwords; and at the pragmatic level.

### A. *Limitations at the Semantic Level: Emoji and Symbols*

The first limitation of the auto-generated clustering was that visual information such as emojis and symbols were excluded from the analysis targets. A comparison of clusters suggested that such visual information may play a crucial role in modulating the literal meaning of a word. For example, Rinna phrases that indicated "yes" included variations such as "yes ♡" and "ah, yes." Additionally, unspoken implications were expressed by emojis and symbols attached to phrases.

TABLE III
AUTO-GENERATED CLUSTERING OF THANK-CLASS PHRASES

| Cluster index | Phrases | Preprocessed Phrases | Tokenized Phrases |
|---|---|---|---|
| Cluster A | ありがとう☻ | ありがとう | [ありがとう] |
| | ありがとうございます！ | ありがとうございます | [ありがとう] |
| | ありがとう！ | ありがとう | [ありがとう] |
| | ありがとう | ありがとう | [ありがとう] |
| Cluster B | 三┗(┓+^o^)+ドゥルルル | 三++ドゥルルル | [+, +, ドゥルルル] |
| | + | + | [+] |
| | ありがとー♪ | ありがとー | [ありがと] |
| Cluster C | あざます | あざます | [あざ] |
| | あざます！！！ | あざます | [あざ] |
| | あざっし‼ | あざっし | [あざ] |
| | あざます😭 | あざます | [あざ] |
| | あざます！！ | あざます | [あざ] |
| | あざます🙌🙌 | あざます | [あざ] |
| | あざます🙏💦 | あざます | [あざ] |
| | あざます💕 | あざます | [あざ] |
| | あ ざ ま す 😍😍😍 | あざます | [あざ] |

a. The “+” symbol represents Manji, which is defined in the Unicode code-point of U+534D.
b. ありがとう, ありがと, and あざ indicate “thank you.”
c. Only representative phrases are shown for each cluster.

TABLE IV
MANUAL CLUSTERING OF THANK-CLASS PHRASES

| Nameplate | Phrases |
|---|---|
| Thank | ありがとう☻/ ありがとー♪ / あざっし‼ /<br>ありがとうございます！ / ありがとー(((o(*ﾟ▽ﾟ*)o))) /<br>あざっし。/ あざます / あざます！ / あざます！！ /<br>あざます！！！！！！！！！！ / あざます👍 /<br>あざます👍👍 / あざます💕 / あざます😍😍😍 /<br>あざます😭 / あざます🙌🙌🙌 / あざます🙏💦 /<br>やだ、ありがとう😫💔 / ありがと / あざます！！！ |

a. ありがとう, ありがと, and あざ indicate “thank you.”
b. Only representative phrases are shown for each cluster.

With manual clustering, the yes-class phrases were divided into three clusters: “yes (positive),” “yes (reluctantly),” and “yes (laugh)” (Table I). In contrast, the auto-generated clustering placed these phrases into a single cluster (Table II). The Rinna phrases that shared a literal meaning were divided into several clusters depending on their nuances. Considering the cheerful and expressive character of Rinna, a concept should be prepared for each nuance. Therefore, we concluded that clustering should consider not only the literal meaning of words, but also emojis and symbols.

### B. Limitations at the Semantic Level: Slang, New Words, and Buzzwords

The second limitation of the auto-generated clustering was the difficulty in reflecting vocabulary outside standard idioms.

TABLE V
MANUAL CLUSTERING INCLUDING MANJI

| Nameplate | Phrases |
|---|---|
| Awesome | + / まじ+ / 三┗(┓+^o^)+ドゥルルル<br>Σd(・ω・d)ｵｩｨｪ!!! / 最高 / ٩( ^o^)╮-=ﾆ= =三(ɜ ̈ﾊﾟｧ)<br>/<br>(*´ω`*) / (´▽`)ノ♡♡ / (*´︶`*)ほほ / (*′皿`艸) /<br>(*-`ω´-) 9 ﾖｯｼｬｧ!! / v(. ・・. )ｲｴｯ♪ / 万次 /<br>(っ'ヮ'c)♥← / (* ˋ・ω・ˊ*) ﾝ うぃ☆ / (☝ ՞ਊ ՞)☝ｳｪｰｲ |

a. The “+” symbol represents Manji, which is defined in the Unicode code-point of U+534D.
b. 最高 indicates “awesome.”
c. + and 万次 are both pronounced as “manji,” indicating the same usage.
d. “ﾖｯｼｬｧ,” “ｵｩｨｪ,” and “ｳｪｰｲ” indicate emotional excitement.
e. Only representative phrases are shown for each cluster.

TABLE VI
AUTO-GENERATED CLUSTERING INCLUDING MANJI

| Cluster index | Phrases | Preprocessed Phrases | Tokenized Phrases |
|---|---|---|---|
| Cluster A | お、おう | おおう | [おおう] |
| | まじ+ | まじ+ | [まじ, +] |
| | ちゅちゅちゅ | ちゅちゅちゅ | [ちる, ゅちゅちゅ] |
| Cluster B | 三┗(┓+^o^)+ドゥルルル | 三++ドゥルルル | [+, +, ドゥルルル] |
| | + | + | [+] |
| | ありがとー♪ | ありがとー | [ありがと] |

a. The “+” symbol represents Manji, which is defined in the Unicode code-point of U+534D.
b. ありがと indicates “thank you.”

Although colloquial idioms (e.g., new words and buzzwords) and standard idioms share the same concept, they may be composed of different words. In such cases, auto-generated clustering may divide colloquial and standard idioms into different clusters, which is not ideal. For example, Rinna phrases that indicate “thank you” include variations such as ありがとう☻ (Arigatou☻), ありがとー♪ (Arigato–♪), and あざっし‼ (Azasshi!!). Although all of these phrases share the same concept of “thank you,” the auto-generated clustering divided them into different clusters (Table III). This division was likely caused by differences in the words comprising the phrases (i.e., “tokenized phrases” in Table III). In particular, あざ (Aza) is a new type of internet slang that means “thank you.” Unlike the auto-generated clustering, the manual clustering grouped these phrases into a single cluster (Table IV).

Similar to slang, it was difficult to cluster multiple buzzwords using auto-generated clustering. For example, Manji is a rarely used Japanese character that has recently become a buzzword among junior high and high school students (see the Unicode code-point of U+534D for the exact representation). Although Manji was originally used to indicate symbolic objects such as a family crest, it has recently been used to express emotional excitement or something extreme. Therefore, a phrase that contains Manji should be clustered with phrases such as 最高 (Saikou meaning

TABLE VII

MANUAL CLUSTERING OF PHRASES INCLUDING *II KARA*

| Nameplate | Phrases |
|---|---|
| Reject | いいから。<br>いいから<br>o(*≧□≦)o ダメ～！！<br>ほんとむり！ |

a. ダメ and むり indicate "no."

TABLE VIII

AUTO-GENERATED CLUSTERING OF PHRASES INCLUDING *II KARA*

| Phrases | Preprocessed Phrases | Tokenized Phrases |
|---|---|---|
| いいから | いいから | [いい] |
| いいから。 | いいから | [いい] |
| いいよ | いいよ | [いい] |

a. いい indicates good.

"awesome") and (´▽`)ノ♡♡ (Table V). However, the auto-generated clustering divided phrases containing Manji into two clusters (Table VI), making it difficult to find a shared concept inside each cluster. This failure to cluster Manji may have been caused by failed mapping. Because Word2Vec was trained on Japanese Wikipedia articles, its mapping was unlikely to reflect the recent usage of Manji. Therefore, we concluded that clustering should consider slang, new words, and buzzwords.

### C. *Limitations at the Pragmatic Level*

The third limitation of the auto-generated clustering was the difficulty in reflecting new usages of standard words. For example, the phrase いいから (ii-kara) can reflect both positive and negative nuances depending on its usage. Ii-kara consists of two standard words: ii (good) and kara. Because kara is a conjunctive particle without an independent meaning, ii-kara may have a positive nuance (i.e., good). However, in recent years, kara has been used to convey a negative emotion when used at the end of a phrase [38]. Therefore, ii-kara is often used with negative nuances, such as rebuttal and irritation. Considering the recent usage of kara, ii-kara should be clustered with words of refusal such as o(*≧□≦)o ダメ～！！ (o(*≧□≦)o Dame–!!) meaning "Noooo!!" and ほんとむり！ (Honto muri!) meaning "Seriously, it's impossible!' (Table VII). However, the auto-generated clustering grouped ii-kara into a "good" cluster (Table VIII). Therefore, we concluded that clustering should consider new uses of standard words.

## V. FUTURE STUDIES FOR DEVELOPING A PRACTICAL GESTURE GENERATION SYSTEM.

The main objective of our study was to identify the difficulties in obtaining concepts for a unique vocabulary personalized for a conversational agent (Fig. 1). A comparison between manual and auto-generated clustering identified three limitations of the NLP-based approach: at the semantic level with emojis and symbols; at the semantic level with slang, new words, and buzzwords; and at the pragmatic level. We attribute these limitations to the personalized vocabulary of Rinna.

Based on the results, this section discusses directions for future study on incorporating both NLP-based techniques and manual clustering methods into the implementation of the concept-based approach.

### A. *Strategy for Leveraging Manual Clustering*

Although the experimental results demonstrated the effectiveness of employing human clustering to obtain concepts, manual clustering is labor intensive, particularly when dealing with big data, such as the larger phrase sets of Rinna. One possible solution to this scalability problem is a hybrid method that incorporates both manual clustering and NLP techniques.

- Based on the manual clustering of reasonably small-sized phrases, one can obtain an agent-specific concept set (i.e., nameplates) and several representative phrases (i.e., labels) to be included in each concept.
- By preparing an appropriate text-context metric, an input phrase of an agent can be mapped to one of the defined concepts. As an example of the metric, an input phrase can be mapped to a concept that contains the nameplate closest to the input in an NLP-based embedding space.

Such a hybrid method would enable manual clustering to be scaled to a larger phrase set while retaining concepts that reflect the personality of an agent.

### B. *Strategy for Improving NLP-Based Approaches*

The limitations of NLP-based techniques identified in this study still need to be addressed, even if a hybrid method is adopted. The Rinna vocabulary is characterized by the use of emojis and kaomoji. These symbols allow short sentences to incorporate various emotions. Additionally, the use of these symbols can reduce the psychological distance between a user and a conversational agent. In 2014, emojis were standardized in Unicode, making it easier to install them in various systems [39]. Therefore, we can reasonably assume that the use of emojis or kaomoji will become common for conversational agents in the near future.

To handle these symbols using an NLP-based technique, a morphological analysis engine or text vectorization method should be trained on a dataset that contains these symbols [40]. Similarly, to cover slang, new words, buzzwords, and new uses of standard words, an NLP-based vectorization method should be trained on a dataset that includes these terms. A candidate dataset could be a Rinna phrase set or text data from a social networking service.

### C. *Methodological Considerations*

We are aware that the manual clustering has two limitations: the small size of analyzed data (i.e., 358 phrases) and that the replicability of results cannot be guaranteed. These limitations highlight the difficulty of manually analyzing data. However, the manual clustering method that we employed was established in the field of cultural anthropology [27]–[29]. Furthermore, the pairing of nameplates with classified Rinna phrases yields reasonable results (Tables I, IV, V, and VII). Therefore, we believe that the concepts obtained by manual clustering reflect the character of Rinna to some extent.

Additionally, because the effect of data preprocessing on the obtained concepts was unknown, we used the same number of clusters identified in manual clustering for auto-generated clustering. Regardless, we believe that our conclusions are not dependent on a particular number of clusters because the key findings regarding the limitations of auto-generated clustering were obtained through qualitative comparisons of all clusters.

## VI. Usefulness Test of the Concept-Based Approach

As this study aimed to identify difficulties in obtaining concepts for a unique personalized vocabulary of a conversational agent, the usefulness of the concept-based approach as a gesture generation system is beyond the scope of this study. However, we conducted a follow-up experiment to understand whether a gesture generation system implementing a concept-based pipeline could output reasonable co-speech gestures for a personalized vocabulary. In this follow-up experiment, we compared human impressions of robot co-speech gestures selected through semantic space mapping to those of randomly selected gestures. Note that this experiment was not intended to evaluate the feasibility of the hybrid method (i.e., a future study) or the effectiveness of manually preparing gesture-concept mappings.

### *A. Methods*

Considering the limitations identified in the auto-generated clustering, we used the manually obtained clusters (i.e., concepts) of the Rinna phrases. The gestures were performed by the Pepper robot from Softbank Robotics. The gestures assigned to the concepts included 205 movements selected from Pepper's preset gestures [41]. The assignment of preset gestures to concepts was performed subjectively by one author, and then confirmed and approved by two other authors. As a result, the number of preset gestures assigned per concept ranged from zero to three. We prepared 10 co-speech gesture videos for evaluation using the following procedure. First, we selected the 10 concepts that appeared most frequently among the Rinna phrases, excluding concepts that could not be assigned a gesture. Next, a phrase-gesture pair was prepared for each concept by randomly selecting a phrase and gesture associated with the concept while excluding phrases containing emojis due to the difficulty of converting them into speech. A gesture video was obtained by recording Pepper performing the gesture. The audio of the phrase being spoken was obtained using a third-party text-to-speech engine TTSMP3 [42]. Finally, co-speech gesture videos were prepared by manually aligning the gesture videos and speech audio to begin simultaneously. As a baseline comparison, 10 shuffled videos were prepared by randomly pairing the videos and the audio.

This experiment was conducted in accordance with the principles of the Declaration of Helsinki and no personal information (e.g., age, gender, or name) was collected. Thirteen volunteers, who were recruited through recruitment campaigns within the authors' organizations, participated in the experiment. The participants were informed of the purpose of the experiment and confirmed that they understood basic Japanese. The evaluation took place online and took about 15 min.

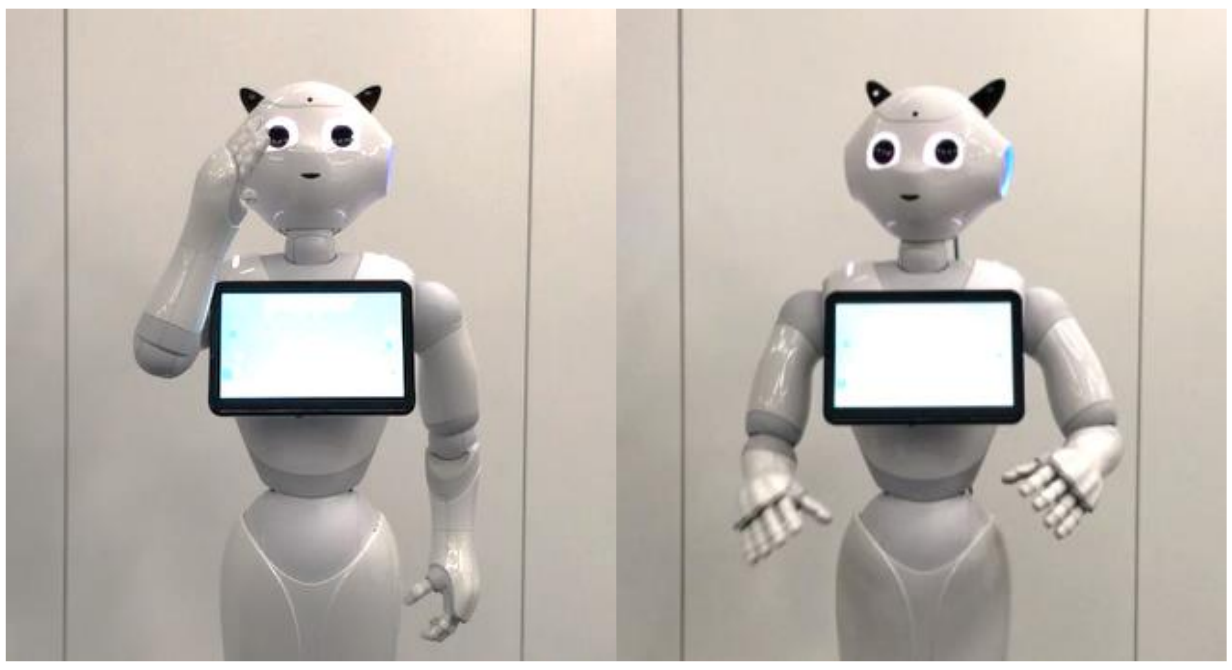

Figure 2. Examples of video clips used for impression evaluation.

TABLE IX
Questionnaire Items Used for Impression Evaluation

| **Questionnaire Items (Japanese)** |
|---|
| (Fake-Natural), (Machinelike-Humanlike), (Unconscious-Conscious), (Artificial-Lifelike), (Moving rigidly-Moving elegantly) |

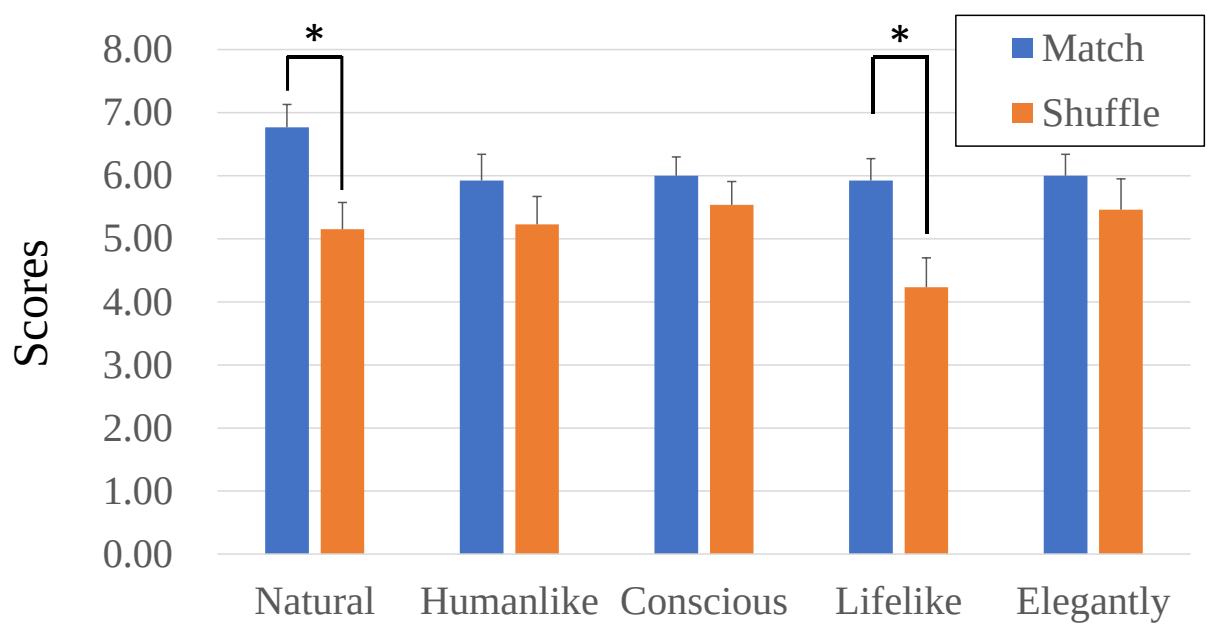

Figure 3. Averages and standard errors of the questionnaire. Statistical differences between matched and shuffled conditions are indicated by markers ($p < 0.05$ with BH correction for 5 contrasts; two-tailed Wilcoxon signed rank test; $N = 13$).

The participants watched four video clips: two contained five videos of the matched pairs and the remaining two contained five videos of the shuffled pairs. Fig. 2 shows an example of a video clip. The order in which the four video clips were presented was shuffled between participants to eliminate the effect of the order. We presented five videos in sequence because we considered that a single stimulus was insufficient for a participant to form an impression of the robot. The number of videos to be presented at a time was decided to simulate the number of turnovers expected in a real conversation session with Rinna while minimizing the risk of participant distraction during a long video. As user input data to Rinna was unavailable, we referred to the report on XiaoIce [43] (i.e., ≥ 5 conversation turns per session), which is a chat engine based on the same Microsoft technology used by Rinna. To maintain the attentiveness of the participants, each time they watched a video clip of five motions, they were asked to select the phrase that the robot said with the first motion using a single-choice question. To avoid distracting participants from the robot gestures, subtitles were not provided.

A questionnaire was used to measure an anthropomorphic index based on the Godspeed survey [44]. The index included five questions (Table IX). The participants answered the five questions for each video clip based on a five-point Likert scale (e.g., 1 = Fake, 5 = Natural). For the matched and shuffled conditions, the values obtained for each question were summed for both video clips. These summed values were used to evaluate the impressions of each condition. The values for the 13 participants were compared for each question.

### B. *Results*

Fig. 3 summarizes the questionnaire results. There were significant differences in scores for "Natural" and "Lifelike" ($p$ = 0.0176 and 0.0195, respectively; two-tailed Wilcoxon signed rank test, $p$ with BH correction for 5 contrasts < 0.05). This indicates that mapping gestures using a concept-based approach left a better impression than random mapping in terms of anthropomorphism.

Although the scores were generally higher in the matched condition than in the shuffled condition, the differences were smaller for "Humanlike," "Conscious," and "Moving elegantly." The evaluation of "Conscious" may be influenced by speech content in addition to gestures. The differences between conditions may have been ambiguous due to the absence of information regarding the conversations. The questions about "Humanlike" and "Moving elegantly" may have directed the participant attention to the movements of the robot. Because the evaluation of movements is subject to robotic hardware (e.g., appearance and motion smoothness), the differences between conditions may not have been pronounced.

One may think that the random condition was too weak as a baseline. However, according to a study that compared recent gesture generation methods [45], randomly mapped gestures were rated higher by human raters than any of the other methods when compared in terms of the appropriateness of the links between motion and speech. Therefore, we consider random mapping to be a reasonable benchmark. Even though our results merely indicate that the gestures selected by a concept-based pipeline are better than randomly selected gestures, they still suggest the usefulness of a concept-based gesture generation system for a personalized vocabulary.

## VII. Conclusion

In this study, we aimed to identify the issues associated with obtaining concepts for a unique vocabulary personalized for a conversational agent. By comparing the results of manual and auto-generated clustering, we identified three limitations of the latter: at the semantic level with emojis and symbols; at the semantic level with slang, new words, and buzzwords; and at the pragmatic level. We attribute these limitations to the personalized vocabulary of Rinna. Although the results demonstrated the effectiveness of manual clustering, it is impractical because of its labor intensiveness. Therefore, we discuss the hybrid use of manual clustering and NLP techniques as a future research topic for the practical implementation of a concept-based approach. In a follow-up experiment, we demonstrated that gestures selected using a concept-based approach made a better impression than randomly selected gestures for Rinna phrases, particularly in terms of the naturalness and lifelikeness of the robot. These results suggest the usefulness of a concept-based gesture-generation system for personalized vocabulary. We hope that our research will aid in developing a gesture generation system for conversational agents with unique personalized vocabularies. A practical gesture generation system requires additional components, including concept-gesture mapping [46] and ego-noise cancelation during operations [47], [48]. To further our research, we intend to develop a system that integrates these components with phrase-concept mapping based on the hybrid use of manual clustering and NLP techniques.